# OpenCL-based FPGA accelerator for disparity map generation with stereoscopic event cameras

David Castells-Rufas
Universitat Autònoma de Barcelona
Edifici Enginyeria, Campus UAB
08193 Bellaterra, Spain
david.castells@uab.cat

Jordi Carrabina
Universitat Autònoma de Barcelona
Edifici Enginyeria, Campus UAB
08193 Bellaterra, Spain
jordi.carrabina@uab.cat

## ABSTRACT
Although event-based cameras are already commercially available. Vision algorithms based on them are still not common. As a consequence, there are few Hardware Accelerators for them. In this work we present some experiments to create FPGA accelerators for a well-known vision algorithm using event-based cameras. We present a stereo matching algorithm to create a stream of disparity events disparity map and implement several accelerators using the Intel FPGA OpenCL tool-chain. The results show that multiple designs can be easily tested and that a performance speedup of more than 8x can be achieved with simple code transformations.

## 1. INTRODUCTION
Computer Vision has been historically dominated by the analysis of images acquired from frame-based cameras which sense the world by acquiring the light that hits a matrix of photodiodes. The photosensitive cells (pixels) integrate the incoming photonic energy during a period of time known as exposure time. The same period is used to integrate the light for every pixel of the image. If the time is very short or the incident light in a certain area is very low, the acquired value is very low or even zero (which is associated to the black color). On the other hand, if the time is very long, moving objects contribute light to many pixels of the sensor, and regions with strong illumination are saturated to the maximum possible value (corresponding to the white color).

In addition, the readout circuit transfers all the values of the pixel matrix to a computer host. The maximum speed at which this process can be done is defined by equation (1) where $FPS$ is the speed in frames per second, $T_{exp}$ is the exposure time, $BW_{ch}$ is the bandwidth of the communication channel between the camera and the host in bits per second, $bpp$ are the bits used to represent the acquired reading per pixel, and $w$ and $h$ are the width and height of the pixel matrix respectively.

$$\frac{seconds}{frame} = \frac{1}{FPS} = T_{exp} + \frac{(w \cdot h \cdot bpp)}{BW_{ch}} \quad (1)$$

The main drawbacks of this approach are that the common exposure time limits the dynamic range of the camera and that the temporal resolution is limited by the camera speed ($FPS$). To overcome the first problem a possible approach is to work with different exposure times and combine and fuse the acquired information, such as in [1]. Nevertheless, this reduces the effective frame rate and adds complexity to the subsequent algorithms that are typically prepared to work in convenient illumination conditions. The speed limitations are often addressed by increasing the bitrate of the channel, but as market also demands more resolution, which has a quadratic effect on the necessary channel bandwidth, the frame rate hardly increases significantly.

Time-domain imaging groups a number of alternative techniques to frame-based imaging that focus on the temporal evolution of the pixel luminance. A good classification of these alternatives is presented in [2]. Event-based cameras [3] are a subset of such group. Instead of delivering a stream of pulse-code modulated (PCM) pixels at a certain $FPS$ rate, they deliver a stream of bits informing about the pixels affected by significant changes. In some cases (like contrast, or time derivative cameras), just one bit can be used to inform about the increase or decrease of pixel luminance, producing a bitstream that could be interpreted as a pulse density modulated (PDM) signal. Since the information per pixel is lower and it is expected that a significant temporal redundancy exist, the bandwidth requirements are also expected to decrease. The expected spare channel bandwidth can be used to significantly increase the temporal resolution

However, this kind of image sensors also present some challenges. Since pixel events happen asynchronously the readout system must asynchronously deliver each individual pixel with its own coordinate context, increasing the necessary bits per pixel to transmit. Address Event Representation (AER) standard is typically used to transmit this information.

As the temporal resolution is higher and address information increases the information to transmit, the communication channel can be affected by congestion and saturation when many events occur, thus, causing the drop of events.

These constraints produce some challenges for vision algorithms since they must consider asynchronous streams of events at high temporal resolution with some probability of missing information due to channel saturation.

In any case, this can require a lot of computation that typically follows a dataflow model of computation. In many applications, like Advanced Driver-Assistance Systems (ADAS), the total latency of the dataflow pipeline is an important factor as it have safety implications.

We want to design new ADAS algorithms working with this type of cameras and map them to FPGAs to exploit the good performance and latency characteristics of these reconfigurable devices.

In this paper we focus on disparity map generation from event-based stereo cameras. As shown in [4], this problem has already been addressed by many groups using different computing platforms including CPUs, GPUs, and FPGAs. We want to prove the convenience of OpenCL toolchains to design FPGA accelerators for it. One of the expected benefits is a fast design space exploration process.

The paper is organized as follows. In section II we review the state of the art on sensors, algorithms and FPGA implementations. In Section III we describe our proposed process to obtain the disparity map. In Section IV we describe the FPGA implementation. Finally we present results and conclude.

## 2. STATE OF THE ART

Event-based cameras were proposed as bioinspired designs after the advances done in understanding how the human visual system works. A number of academic chip designs have been produced during the last 25 years, and now there are some commercially available chips.

Since they use more transistors per pixel, the resolution and fill factor of such sensors is typically lower than standard image sensors, although this is a drawback that might possibly be overcome in the future with 3D stacked chip design.

Anyhow, the resolution of sensors has been increasing slowly but steadily. One of the first such sensors was presented by Boahen presented in 1996 with a resolution of 64×64 pixels [5]. In 2001, his group increased the resolution to 80×60 [6]. In 2005 Lichtsteiner and Delbruck also created a 64×64 pixel silicon retina with a logarithmic response [7]. In the following year, together with Posch they doubled the resolution to 128×128 [8]. Serrano-Gotarredona and Linares-Barranco also presented a 128×128 sensor in 2013 [9]. A couple of years before, Posch et al. had presented a QVGA (320×240) in [10]. Finally, relatively recent developments in 2017 have gone to VGA (640×480) [11] and WVGA (768x480) [12].

Despite these advances and the already commercial availability of event-based cameras, they are not as widespread as frame-based cameras. Their adoption depends in their ability to prove some superior performance on computer vision applications, but they have to compete with the vast existing research based on frame-based cameras.

To reduce the entry barriers for researchers, manufacturers are also interested in delivering recorded datasets for the analysis of the research community like [13] [14]. With the maturity of virtual reality environments, an increasingly attractive option is to create synthetic data from simulator environments, such as done in ESIM [14] [15].

Computer vision algorithms using event-based cameras is still an open research topic. Some algorithms, like tracking, object detection, gesture detection, etc. have been proposed (see [16, 17]) but results about their superiority are still not conclusive. There is some intuition the Deep-Learning revolution will perfectly fit with the bioinspired nature of event-based sensors making them generally more adequate than frame-based ones, but, to the best of our knowledge, this still remains unproven.

Stereo matching is one of the algorithms that has been studied with some depth. In a setup with two cameras the algorithm tries to find the related events on both cameras so that scene depth can be inferred. Figure 1 depicts an ideal case where a number of events happen over a short period of time. For simplicity we ignore the y coordinate and just display the x coordinate of the events. Each event is always matched for a corresponding event in the other sensor, which happens at the very same instant. From the disparity in the horizontal dimension we can (shown as a black arrow in the image) we can deduce the depth of the event in the scene.

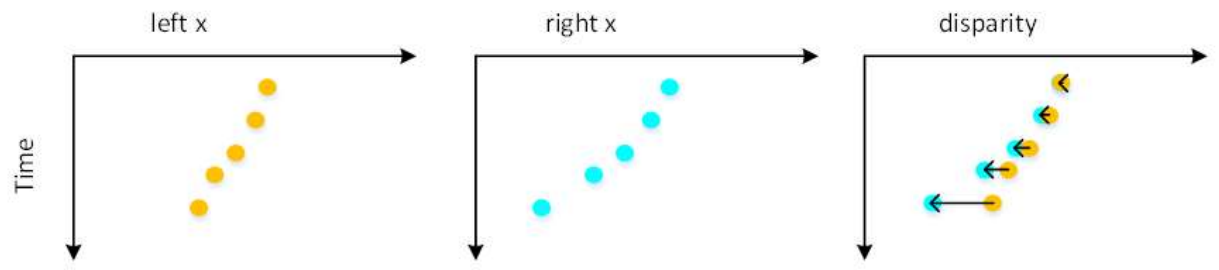

**Figure 1 ideal scenario for matching events from left and right cameras. The y coordinate of events is ignored. Left) events from the left camera over time. Center) events from the right camera over time. Right) matching of events, and disparity calculation.**

This ideal case is obviously far from what happens. In reality the scenario is much more complex. Multiple events happen simultaneously so finding the best match for an event is almost impossible if just the information of one event is used. This is especially true if the event information is a 1 bit polarity. Moreover the real corresponding event can be sensed in a slightly different instant or it can never be sensed. To make it harder false events can be triggered as result of noise.

Individual events do not have enough unique information to allow their identification and match. Most algorithms need to define spatial and temporal constraints to the matching candidates, or create features from multiple events and try to match the features instead of the individual events.

Rogister et al. [18] work on individual events. They limit the potential matches to the events that have a similarity in the number of positive and negative events along the epipolar line and happened in a similar time window. However, the existing many ambiguities and the jitter of the event timestamps limit the performance of the algorithm.

Carneiro et al. [19] also work at the event level. Their approach tries to improve the performance by applying a Bayesian inference model with a multiview setup with up to 6 cameras. This increases the complexity of the system setup, and the results are improved but jitter is still a cause of a low success rate.

Xie et al. [20] present a method which uses a similar approach to find the best event candidates for a match, but later apply a belief propagation smoothing so that the result has more coherency. This approach remove some of the ambiguities by including a consistency constraint with neighboring pixels.

A different family of approaches try to combine information of groups of events to create more unique features that can be more easily matched. One early such approaches was proposed by Kogler et al. [21]. They compose a standard frame from the event stream and then compute the disparity map with classical frame based algorithms. The integration is done by integration the events over a 10 ms window. In [22] , Schraml et al. follow a similar approach but analyze a number of region similarity operations.

Most algorithms try to minimize the use of memory resources, which is considered desirable as being more close to bioinspired neuromorphic circuits. Camuñas et al. [23] combine the information from groups of pixels by applying Gabor Filters at various scales with few intermediate memories. Adreopoulos [4] removes the use of temporal memories by morphological operators and Hadamard product on features with various scales.

Regarding implementations of event-based vision algorithms based on FPGAs, there are some precedents in the literature such as robot arm control [24], rotation speed identification [25], optical flow [26] [27] and stereo matching [23] [19] [28]. All analyzed implementations are coded in HDL languages such as VHDL and Verilog.

To the best of our knowledge this is the first attempt to implement FPGA accelerators for event-based vision algorithms using OpenCL.

# 3. STEREO MATCHING

In this work we want to explore some algorithmic ideas and their fast implementation in FPGA accelerators. One of the challenge for this process is to get meaningful data that can be used in a systematic way.

## 3.1 Generating Stereo Camera Streams

To be able to compare the performance of different methods it is necessary to use the same input data for all tested algorithms. So, although the final goal can be to include the algorithm in an embedded system working in real-time with the sensor data, during the development phase it is necessary to work with recorded event streams.

An option was to record the events coming from a Prophesee stereo sensor. Other options might be using existing datasets [13] [14] or using a simulator to generate event streams [15].

The use of a real hardware for algorithm exploration has the drawback of lack of depth ground-truth unless a complex setup combining other sensors is built.

With existing datasets the drawback is that recordings are limited, and they might put the focus on scenarios that you are not targeting. In our case we are targeting a scenario with a fixed position camera sensing moving objects on the scene. The simulation approach is more convenient as it can be controlled at will and provide depth ground-truth. Instead of using ESIM [15] we implemented a much more simple simulator using the JMonkey 3D Engine [29] to produce events from an virtual stereo camera. The simulator is available online at https://github.com/davidcastells/DVSSimulator

Its process loop renders the 3D scene and captures a frame from both virtual cameras, that are separated by a certain baseline. Originally, the response of the physics module and rendering engine of the JMonkey framework is controlled in real-time to match the system's clock. The framework tries to get the maximum number of frames while ensuring that virtual and real time are synchronized. We break this synchronization to make sure that we render frames in a microsecond resolution (in virtual time), which is the typical order of magnitude of temporal resolutions found in event-based cameras.

Last rendered frames are stored so that the temporal difference between frames can be computed. Due to the high temporal resolution the differences between frames are small. Only the pixels with some difference will trigger polarity events. Since polarity events just support 1 bit value increments or decrements, larger differences produce a burst of events.

We produce two streams of data, one for each virtual sensor. We ensure that events are correctly timestamped and are delivered in order. The timestamp jitter (frequent in real sensors) is here inexistent.

## 3.2 Stereo Matching Method

Our simulator produces streams of higher quality that real cameras. Real event-based stereo camera streams may miss events matching corresponding events of the other sensor. Some of those misses could be caused by noise, by channel saturation or other artifacts. Nevertheless, this can also happen when using simulators due to occlusions and different perspective. In such cases events visible by a sensor cannot be visible for the other (see Figure 2).

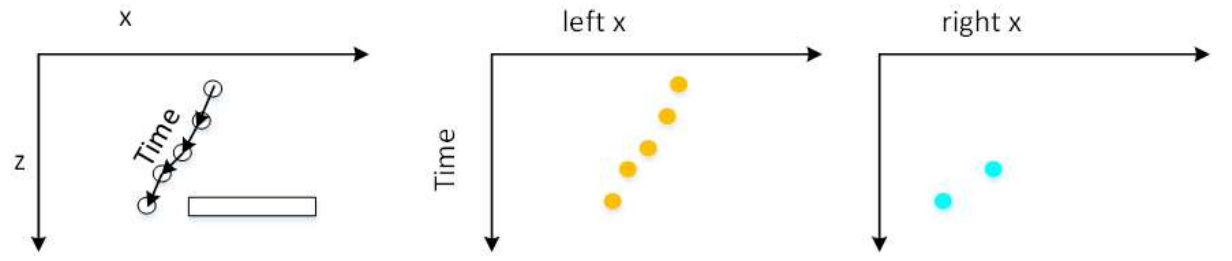

**Figure 2 Occlusions can be a source of missing events. Left) Evolution of the location of an object along time. Middle) Events captured by Left sensor as time progresses. Right) Events captured by Right sensor. Since the events happen behind the rectangular box which blocks the line of sight from the right sensor, it misses those events.**

These effects are important when trying to match events from both sensors. As seen in previous section, algorithms based on trying to match individual events perform worse to those combining information from several ones.

We take a simple approach similar to Kogler [21]. We denote as $P_{x,y}(t)$ as the time sequence of events at location x, y. In the first step of our approach, we integrate polarity change events that happen in a similar time frame, from $t_i$ to $t_f$ (see Eq. 2).

$$A_{x,y}(t) = \sum_{t \in [t_i, t_f]} P_{x,y}(t) \quad (2)$$

This *polarity aggregation* block takes incoming events and stores them in a temporal buffer indexed by location. If there was no event for the event location it only stores the new event, but if the location was already occupied its polarity value is incremented and its timestamp is updated to the last one. Each pixel position has a deadline for "inactivity" (*d*). After this deadline if the pixel has no more activity it is removed from the buffer generating an event which is passed to the next block. Figure 3 depicts this process.

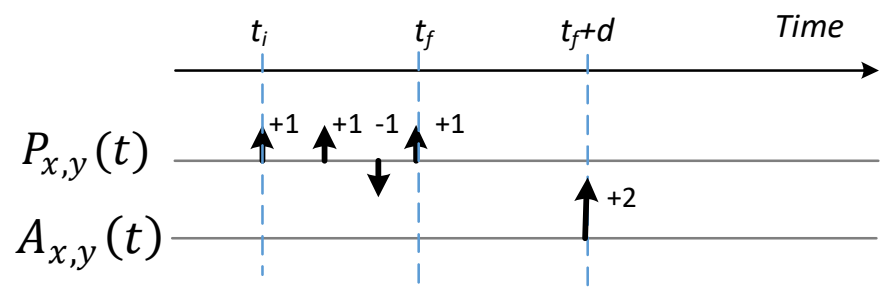

**Figure 3 Polarity aggregation**

By aggregating the polarity events we aim to reduce the number of firings that will trigger the following blocks of the dataflow. Aggregated polarity events $A_{x,y}(t_i)$ are passed to the *level producer*, which only integrates them into a frame $L_{x,y}(t_i)$ and triggers an event with the integrated value to the next disparity map module for each received event.

$$L_{x,y}(t_i) = A_{x,y}(t_i) + L_{x,y}(t_j) \quad (3)$$

Each *level producer* event triggers a disparity map computation. The candidate pixels for the events are only taken from the epipolar line. The disparity value is only computed for the location of the event, all other frame locations are not computed. This avoids unnecessary computation, following a principle similar to the used in Schraml [22]. In our case we use the Sum of Absolute differences of the regions around the event location and the disparity candidate points as the dissimilarity operation (see Eq. 4). The final disparity $d$ is the value that minimizes the SAD for the range of tested values.

.

$$SAD(x, y, d) = \sum_{k_x=-B}^{B} \sum_{k_y=-B}^{B} \left| L^L_{x+k_x, y+k_y} - L^R_{x-d+k_x, y+k_y} \right| \quad (4)$$

The overall process is depicted in Figure 4. The events from left and right cameras are passed to polarity aggregators, then to level producers and finally combined by the disparity map module. Besides the dataflow nature of the process each module contains intermediate memories needed for their processing steps.

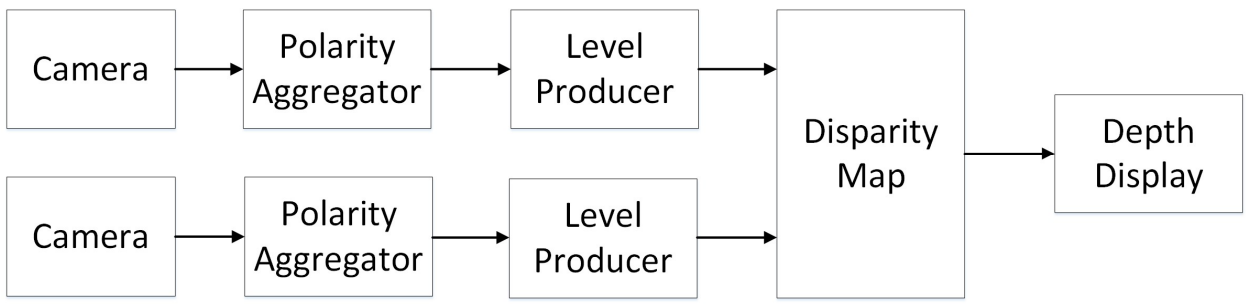

**Figure 4 Diagram of the whole data flow approach.**

In our case, the result is shown on a display. But it could be used as an input for more complex applications, such as obstacle detection and collision avoidance systems.

# 4. FPGA ACCELERATOR OPENCL DESIGN

The use of custom hardware, or application specific circuits implemented in FPGAs to accelerate an existing host computer application has been studied extensively [30, 31]. In the early days the term coprocessor was more used than accelerator, but accelerator was later more accepted (as seen in Figure 5).

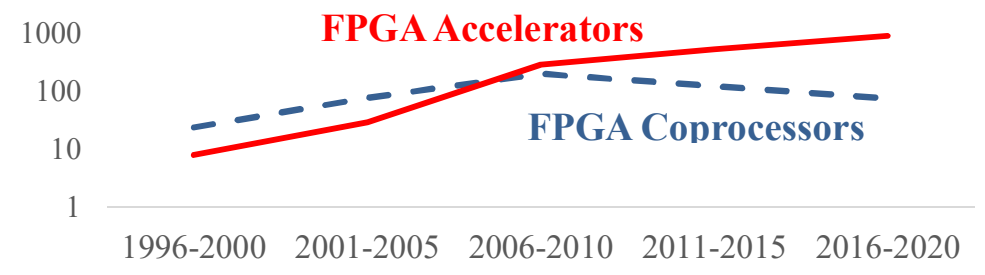

**Figure 5 Evolution of the number of papers using the terms "FPGA Coprocessors" and "FPGA Accelerators" in google scholar from 1996 to 2019 (vertical scale is logarithmic)**

FPGAs are typically programmed with Hardware Description Languages (HLD) such as Verilog and VHDL, which usually use lower levels of abstraction to describe circuits but are not very productive. Higher Level Synthesis (HLS) tools were proposed to create hardware descriptions from software descriptions with directive annotations to increase the design productivity.

There are several proposed methods, but the OpenCL has been adopted by the leading manufacturers and is one of the most populars to implement accelerators [32].

In this paper we implement some designs using starting from a C/C++ and using the Intel OpenCL toolchain targeting the Terasic DE5Net PCIe board.

The central point of OpenCL implementations are the kernels, which are small pieces of C/C++ code with potential high parallelism that are executed in the accelerator. In the OpenCL model there is a memory region in the accelerator device where the host typically uses to transfer the input data and collect the results of the kernel execution.

## 4.1 Original C/C++ Version

In our initial software implementation we use a modular object oriented approach. Each functional block shown in Figure 4 is implemented in a class.

In our case the cameras are replaced by parsers of the recorded streams. The invocation of the modules follows a push fashion. The main application loop consist of getting events from the input streams and pushes them to the next module and triggers its processing function.

The Disparity Map module is invocated by the left and right processing flows. For each side, the levels are stored in the internal buffer, but the stereo match is only performed when receiving left camera events.

The final computed disparity events are shown in an XWindow display for verification. In the rest of the paper we will consider the rate of events arrival to the display module as the metric to optimize, as it is an indicator of the speedup achieved thanks to the accelerator. After compiler optimization, the initial software version is able to display disparity events at a rate of 50 kev/s.

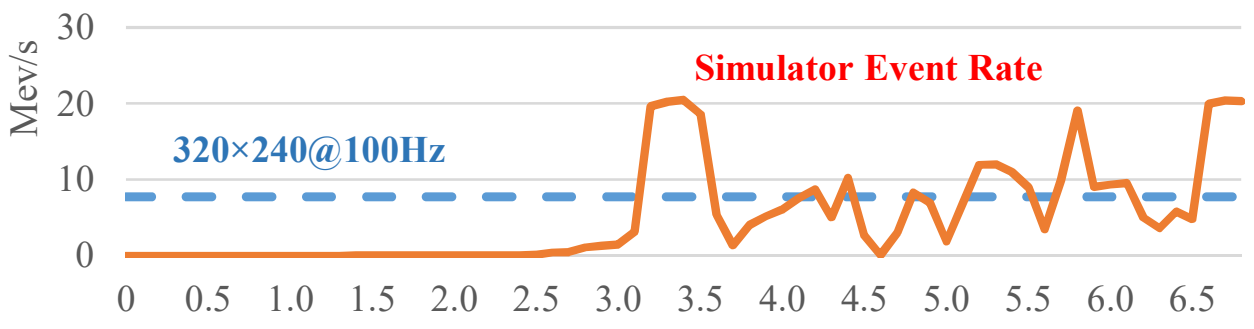

**Figure 6 Evolution of the production rate of events along time. Horizontal axis is time and Vertical axis is Millions of events per second. The orange line shows the number of events generated by the simulator with our input virtual sequence. To put it in context, we add a blue dashed line showing the number of events that would be generated by a QVGA sensor working at 100 FPS.**

This value is not enough for a real-time processing of the stream. As expected (and shown in Figure 6), simulator events have a bursty distribution. The number of generated events depends on the activity in the scene, the minimum delta time $\delta_{min}$ from consecutive events in the same pixel, and the maximum possible luminance increment $\Delta_{max}$, since big increments are translated into a train of events.

$$ER_{max} = \frac{wh\Delta_{max}}{\delta_{min}} \quad (5)$$

The worst case will be given by the case of having activity in all the pixels of the scene. In that case, the worst case event rate $ER_{max}$ will be determined by Eq. (5). Although it can be extremely large, we consider that the algorithms should support event rates in the order of millions of events per second to justify the use of this kind of devices.

## 4.2 Simple OpenCL Kernel

A simple OpenCL kernel implementation consists on substituting the original polarity aggregator by a wrapper that sends the information to the kernel with almost the same code. Figure 7 depicts the architecture of such implementation.

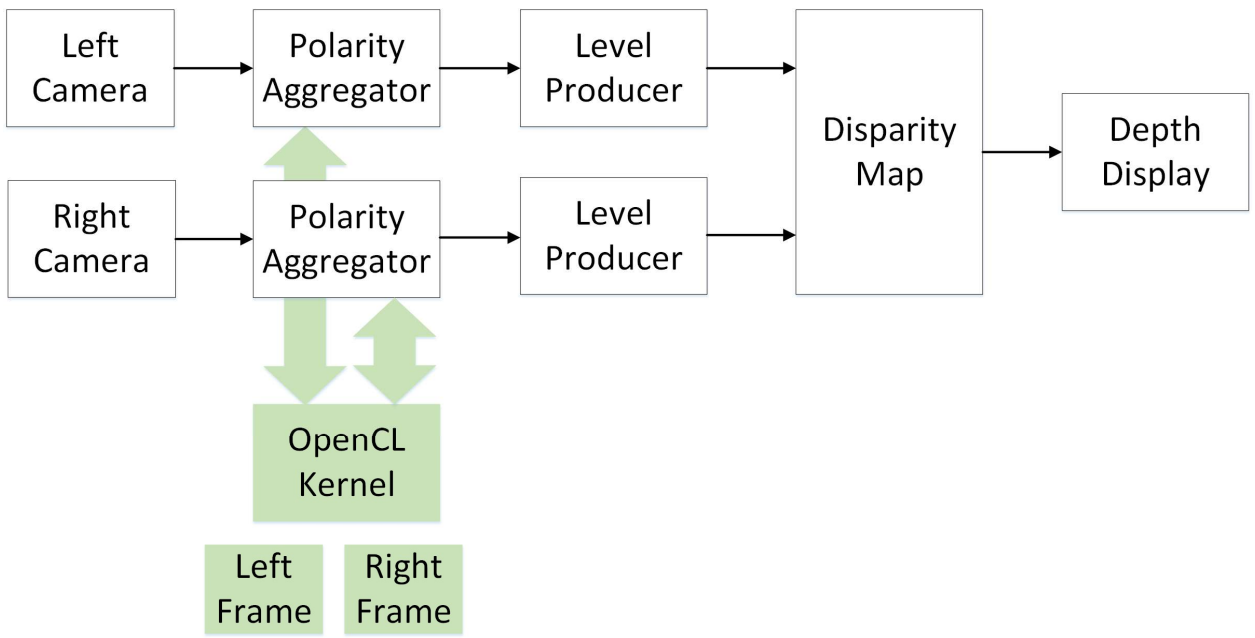

**Figure 7 Application diagram with a simple Kernel invocation from Polarity Aggregators.**

A fundamental difference between object oriented programming languages (like C/C++) and Hardware implementation is the notion of the dynamic instantiation of objects in the Heap. In our software implementation, polarity aggregator objects use a dynamically allocated private buffer to integrate the incoming polarity values. In Hardware we cannot dynamically allocate objects, so the corresponding left and right frame memories must be statically allocated on the OpenCL code (see lines 8 and 9 in the source code below). This also requires explicit initialization of the memory contents. So, in the first kernel invocation an additional parameter is passed to reset the contents of the private memories (see line 10).

The same OpenCL kernel is used for both aggregators. When the host invokes the kernel it uses the parameter *src* to determine which object is referring to.

```
1 __kernel void polarityAggregatorOpenCLBlock(
2  __global int* restrict buffer,
3  int size, int aggregationTime,
4  __global int* restrict out,
5  int scr, int initialize)
6 {
7     ...
8     int rpol[NUM_SCR*IMG_W*IMG_H];
9     int rtime[NUM_SCR*IMGE_W*IMG_H];
10    if (initialize == 1)
11        for (int i=0; i<(NUM_SCR*IMG_W*IMG_H); i++){
12            rpol[i] = 0; rtime[i] = INT_MAX;
13        }
14    for (int i=0; i < size; i++) {
15        ts = buffer[i*4+0]; x = buffer[i*4+1];
16        y = buffer[i*4+2]; pol = buffer[i*4+3];
17        thr = timestamp - aggregationTime;
18        lpe = onEvent(scr, ts, x, y, pol,
19                thr, out, lpe, rpol, rtime);
20     }
21     out[lpe*4+0]= -1;
22}
23
24 int onEvent(..., int thr, __global int* out,
25 int lpe, int rpol[], int rtime[])
26 {
27  ...
28  for (...)
29   for (...)
30      if (rtime[...] < thr) {
31        out[...]=...
32       }
33  ...
34 }
```

In the original software implementation events were individually processed by the dataflow modules. Their processing by the OpenCL kernel must be grouped in packets of events. Otherwise the overhead of the transmission to the FPGA would limit the performance of the system. Nevertheless, this packetization increases the latency of the system.

The kernel inspects all the events in the input event packet and process them by invocating the *onEvent* function (see line 18). In this function the memory is checked to see if the deadline has been met for every pixel. If so, it triggers an event, writing to the out buffer. When the whole packet has been processed the value -1 is written to the output buffer to signal the last valid output value (see line 21).

The invocation from the host is simple. We create an *NDRange* of just 1 *workitem*, transmit a packet of data from the stream, execute the kernel, and capture the result. Single worktime kernels are implemented as pipelined designs by the Intel OpenCL toolchain.

## 4.3 Host Pipes

The problem with the former kernel is that the mechanism to transfer the input data limits the performance of the system. Both, throughput and latency could be reduced by using alternative approaches. We could try to minimize the latency caused by packetization by using transparent communication channels between the host application and the kernel.

This technique was recently presented by Kang and Yiannacouras in [33]. When using it, the kernel receives two channels as parameters (see line 3 and 5 on the code below) . Input and output data is transferred through them instead of input and output buffers.

```
1 __kernel void polarityAggregatorOpenCLPipe(
2   __attribute__((intel_host_accessible, blocking))
3   __read_only pipe int4 host_in,
4   __attribute__((intel_host_accessible, blocking))
5   __write_only pipe int4 device_out,
6   int aggregationTime)
7 {
8  ....
9  int4 indata;
10 while (1) {
11    if (read_pipe(host_in, &indata) == 0) {
12        ...
13        onEvent(...);
14    }
15 }
16}
17int onEvent(..., int thr,
18  __attribute__((intel_host_accessible, blocking))
19  __write_only pipe int4 ch_out,
20  int rpol[], int rtime[])
21 {
22  int4 od;
23  ...
24  for (...)
25   for (...)
26     if (rtime[...] < thr) {
27     ...
28     write_pipe(ch_out, &od);
29     ...
30    }
31  ...
32 }
```

This type of kernels usually consist of an infinite loop that continuously fetch data from the input channel and process them. In our case we use an *int4* channel (see line 9), which is understood as a channel that transmit packets of 4 *int* values. We code all event information in 4 *ints* and read them from the channel (line 11) before calling the *onEvent* function.

On the other hand, instead of writing to an output buffer we write to the output channel (line 28).

Previous kernel was invoked continuously, and each invocation used the parameter *src* to indicate whether right or left camera events were being processed. In this case, the kernel is only invoked once and the events are pushed into the channels. This requires that the screen reference (either left or right) must be passed by the host in every event.

We coded the kernel, but unfortunately the board that we are using does not support the host pipes.

## 4.4 Minimization of memory transactions

Another option to reduce latency and increase throughput is to combine the events of both cameras in the same packet and let the kernel to handle them appropriately. This would reduce the latency by reducing the impact of packetization. The number of invocations to the kernel would be similar, as the total number of events is what drives the number of invocations.

The host application must be modified to create a wrapper that unifies the handling of both cameras events.

In addition, we try to reduce the ammount of memory devoted to each event by packing event information in 3 integers instead of 4.

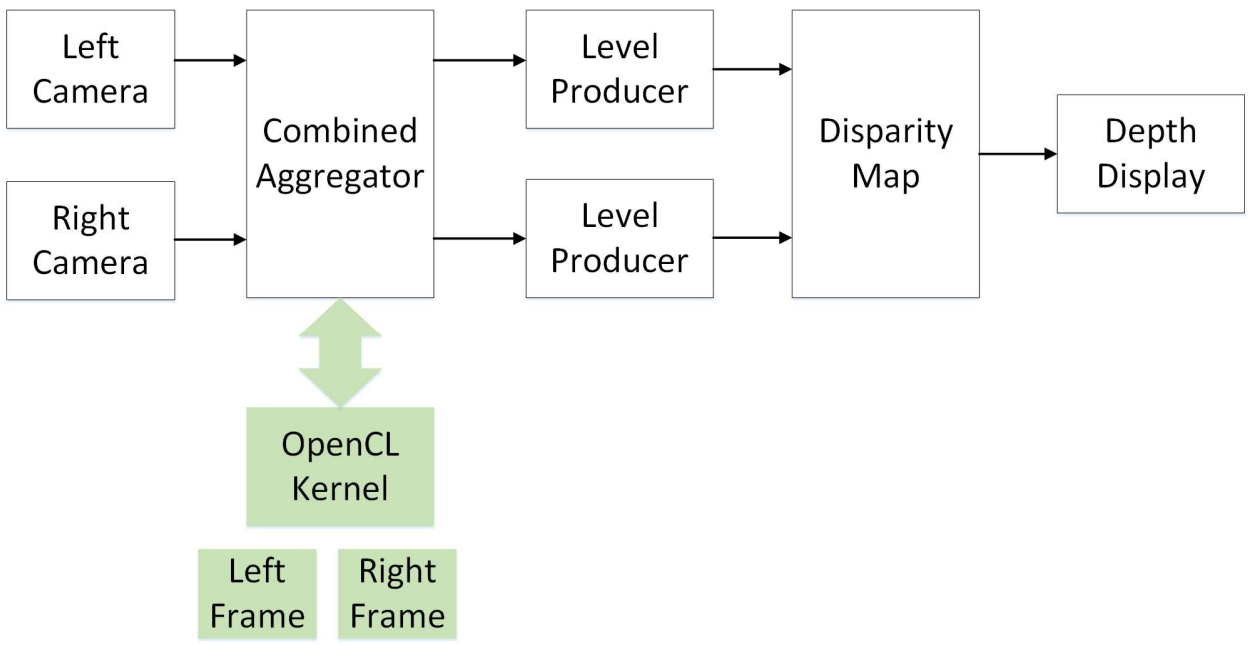

**Figure 8 Application diagram with the wrapper that unifies both event streams to invoke the combined OpenCL kernel**

In this approach the screen parameter (left or right) must be informed in each event, but the kernel source code is very similar to the implementation described in 4.1.

## 4.5 Using Channels

An option to increase the performance of the system is to implement more modules of the original software in the FPGA. An option could be using standard memory buffers to exchange the data between them and having the host to orchestrate their communication. This approach requires that the host coordinate the memory transactions and kernel invocations.

Another option is to use direct OpenCL channels between kernels, as done in [34]. Channels are implemented using Hardware resources transparently and minimize the use of global memory, reducing the latency and increasing the bandwidth between the kernels.

The application must be modified to use a new module that wraps the invocation of two kernels (see Figure 9). The first kernel will be invoked passing an array of the combined events from left and right cameras and will produce no output. Instead, its output will be written to a communication channel shared with the second kernel. The second kernel will read from the channel, process the events and write the result to the output buffers that were passed as a parameter to the second kernel.

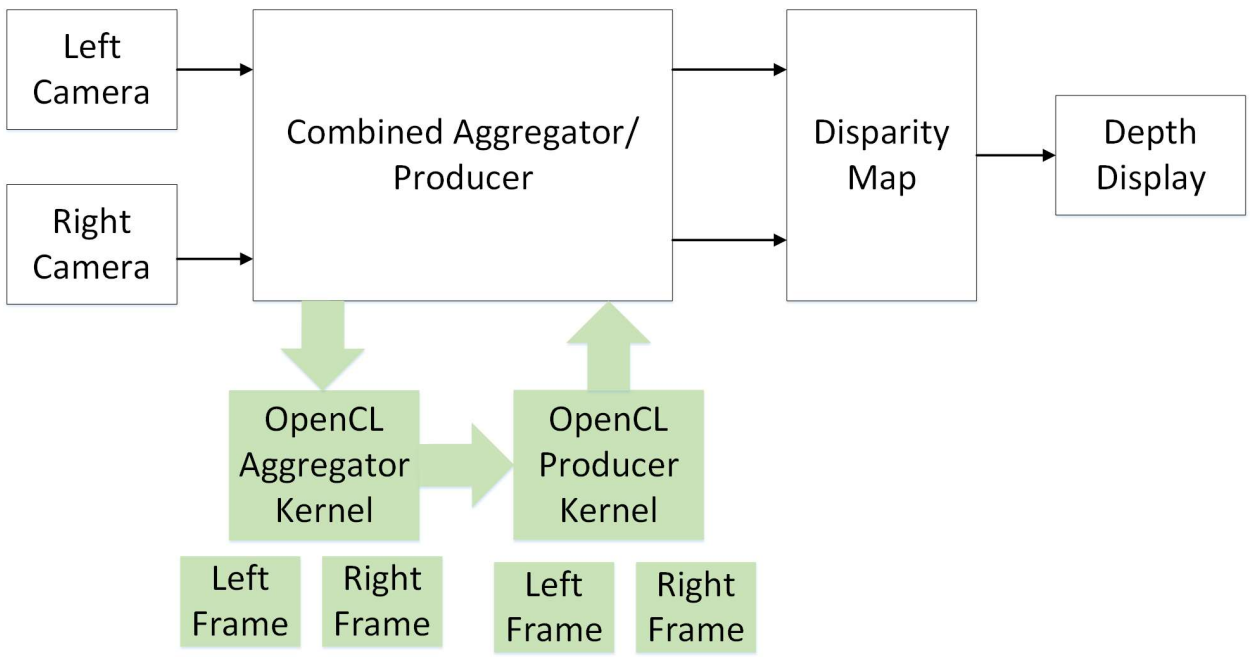

**Figure 9 Application diagram with the wrapper of the channelized kernels**

Unlike in the host pipes case (see section 4.3), the channel is not an input parameter of the kernel function but a global variable statically defined (see line 1 in the code below).

The communication between both kernels is simple. The first kernel writes the processed values to the channel (see line 8) until all events are processed. This event is signaled by writing the value -1 to the channel (line 28). This event is used by the second kernel to know when to stop processing and return the control to the host.

```
1 channel int4 ca2p;
2 void onEvent(...)
3 {
4    ...
5    for (...)
6       for (...) {
7          ...
8          write_channel_intel(ca2p, tx);
9          ...
10       }
11    ...
12 }
13 __kernel void combinedAggregator(
14   __global int* restrict buffer, ...)
15 {
16    ...
17    for (...) {
18       ...
19       onEvent(...);
20    }
21    ...
22    int4 tx;
23    tx[0]=tx[1]=tx[2]=tx[3]=-1;
24    write_channel_intel(ca2p, tx);
25 }
26
27 __kernel void combinedPixelProducer(
28     __global int* restrict out, ..)
29 {
30    ...
31    for (...) {
32      int4 rx;
33       rx = read_channel_intel(ca2p);
34       ...
35       out[1pe*EVENT_STRIDE+0]=ts;
36      ...
37     }
38 }
```

The second kernel receives the values (see line 33), does its processing and writes the resulting values to the output memory buffers.

The algorithm implemented by the second kernel is very simple but requires local memories as well, one for each side (left and right). Thus, the information about the side of events must be embedded in all communicated packets.

## 5. RESULTS

The quality results of the 3D reconstruction are in line with the similar approaches reported in the literature (such as [21]). However, the focus of this work is in demonstrating how OpenCL can be used for the rapid development of FPGA accelerators of event-based vision algorithms.

As reported in [35], OpenCL has proved it suitability as a fast development framework for the implementation of classic image processing algorithms. Although a general claim cannot be done, the achieved performance of some OpenCL accelerators can be comparable or even superior to those implemented with VHDL or Verilog with a significant reduction in the development time.

In our case all kernels presented in this paper were coded in less than a week achieving some significant performance gain. Figure 10 depicts the performance achieved by different implementations. The results are expressed in thousands of events per second. Performance is still lower than the goal of Millions of events.

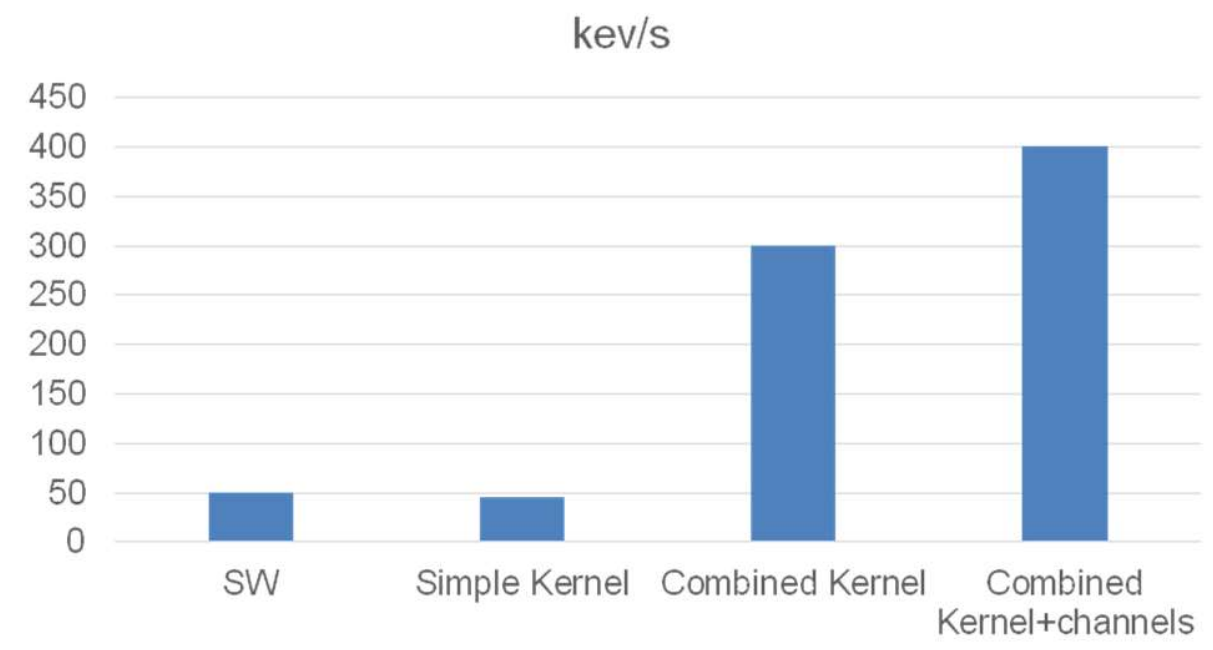

**Figure 10 Performance of the different implementations in thousands of events per second (kev/s) processed by the displaying module.**

The synthesis results of the different kernels are detailed in Table 1. The kernels are executed in a Terasic DE5NET board containing a 5SGXEA7N2F45C2 FPGA device. The synthesis results for the Host Pipes version are not included as they are unavailable on the target platform.

The used resources are around 20% of the logic blocks available in the device. To put the results in perspective consider that the fixed infrastructure needed by the OpenCL runtime already consumes a 10% of the logic resources of the FPGA. Thus, the implementation of the kernels is quite minimal with most resources devoted to unroll the inner loop in the *onEvent* function.

The used memory resources in the different kernels are mainly motivated by the use of local memories to store frame information. In the first *Combined* kernel, frame information is packed in less bytes. This results on less memory usage. On the other hand, the kernel using channels increases the use of memory because the second kernel is also using local memories to store frame information. In this case, the memory resources are not doubled because the amount of local memory used by the second kernel is lower than in the first one.

**Table 1 Synthesis results of different OpenCL Kernels for the 5SGXEA7N2F45C2 FPGA.**

| Design | Perf. | Resources | | | |
|---|---|---|---|---|---|
| | | ALMs | FFs | Memory | DSPs |
| Software | 50 kev/s | - | - | - | - |
| Simple | 45 kev/s | 43 k (19%) | 63 k | 19 Mb (37%) | 0 |
| Combined | 300 kev/s | 44 k (19%) | 65 k | 15 Mb (29%) | 0 |
| Channels | 450 kev/s | 48 k (21 %) | 74 k | 18 Mb (34%) | 0 |

The Figure 11 shows the 3D model generated by the simulator and the disparity map created by the proposed algorithm.

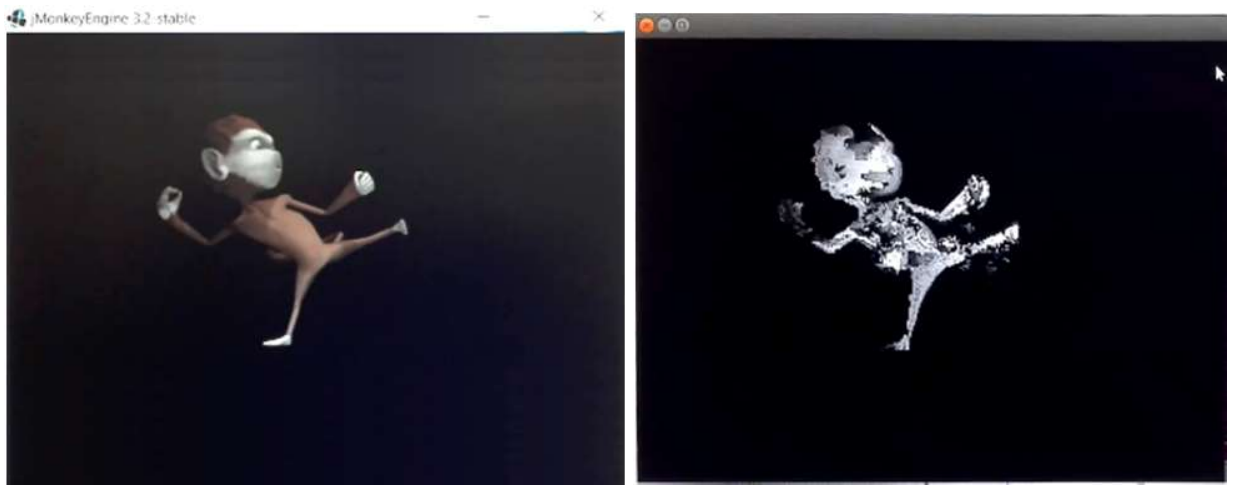

**Figure 11 left) the simulation of the 3D scene. right) disparity map generated by the FPGA accelerated system.**

## 6. CONCLUSIONS

In this paper we proposed a simple disparity map generation algorithm from event-based cameras and we focus on its acceleration on FPGA platforms using the OpenCL programming framework.

As a preliminary step, we implemented a stereo event-based camera simulator to provide data streams that could minimize the jitter and saturation effects found in commercial sensors. The provided streams reach event rates higher than Millions of events per second.

OpenCL is shown as a useful framework for the acceleration of algorithms found in event-based vision algorithms. The maximum speedup factor achieved with our design is 8x with a very short development type of less than a week.

OpenCL provides a flexible platform, with early performance estimators that can ease the programming decisions before investing a long time in Hardware synthesis.

For this type of dataflow applications, we have shown how the use of OpenCL channels can ease the communication between several kernels while reducing the latency and increasing the throughput of the system.

## 7. ACKNOWLEDGMENTS

This project was funded by the Spanish Ministry of Science, Innovation and Universities under the project TEC2014-59679-C2-2-R. We would like to thank Prophesee for their collaboration in providing us an event-based stereo camera and testing datasets.

Our thanks also to Teresa Serrano-Gotarredona from IMSE-CNM for her valuable comments and to Antonio Lopez, from CVC-UAB for his comments and support in FPGA platforms.

## 8. REFERENCES

[1] S. Hrabar, P. Corke and M. Bosse., "High dynamic range stereo vision for outdoor mobile robotics," in *Robotics and Automation, 2009. ICRA'09. IEEE International Conference on. IEEE, 2009..*

[2] D. G. Chen, D. Matolin, A. Bermak and C. Posch, "Pulse-modulation imaging—Review and performance analysis," *IEEE transactions on biomedical circuits and systems,* vol. 5, no. 1, pp. 64-82, 2011.

[3] C. Posch, T. Serrano-Gotarredona, B. Linares-Barranco and T. Delbruck, "Retinomorphic event-based vision sensors: bioinspired cameras with spiking output," *Proceedings of the IEEE,* vol. 102, no. 10, pp. 1470 - 1484, 2014.

[4] A. Andreopoulos, H. J. Kashyap, T. K. Nayak, A. Amir and M. D. Flickner, "A Low Power, High Throughput, Fully Event-Based Stereo System," in *Proceedings of the IEEE/CVF Conference on Computer Vision and Pattern Recognition* , Salt Lake City, UT, USA, 2018. (pp. 7532-7542). DOI: 10.1109/CVPR.2018.00786.

[5] K. Boahen, "Retinomorphic vision systems," in *Proceedings of Fifth International Conference on Microelectronics for Neural Networks. IEEE*, Lausanne, Switzerland, 1996.

[6] E. Culurciello, R. Etienne-Cummings and K. A. Boahen, "High dynamic range, arbitrated address event representation digital imager," *Departmental Papers (BE),* no. 23, 2001.

[7] P. Lichtsteiner and T. Delbruck, "A 64x64 AER logarithmic temporal derivative silicon retina," *PhD Research in Microelectronics and Electronics, IEEE,* pp. 202-205., 2005.

[8] P. Lichtsteiner, C. Posch and T. Delbruck, "A 128 X 128 120db 30mw asynchronous vision sensor that responds to relative intensity change.," in *IEEE Solid-State Circuits Conference, 2006 (ISSCC 2006)*, San Francisco, 2006.

[9] T. Serrano-Gotarredona and B. Linares-Barranco, "A 128× 128 1.5% Contrast Sensitivity 0.9% FPN 3 µs Latency 4 mW Asynchronous Frame-Free Dynamic Vision Sensor Using Transimpedance Preamplifiers," *Journal of Solid-State Circuits,* vol. 48, no. 3, pp. 827-838, 2013. DOI: 10.1109/JSSC.2012.2230553.

[10] C. Posch, D. Matolin and R. Wohlgenannt, "A QVGA 143 dB dynamic range frame-free PWM image sensor with lossless pixel-level video compression and time-domain CDS," *IEEE Journal of Solid-State Circuits* , vol. 46, no. 1, pp. 259-275, 2011.

[11] B. Son, Y. Suh, S. Kim, H. Jung, J.-S. Kim, C. Shin, K. Park, K. Lee, J. Park, J. Woo, Y. Roh, H. Lee, Y. Wang, I. Ovsiannikov and H. Ryu, "A 640×480 dynamic vision sensor with a 9µm pixel and 300Meps address-event representation.," in *IEEE International Solid-State Circuits Conference (ISSCC)*, San Francisco, 2017.

[12] M. Guo, J. Huang and S. Chen, "Live demonstration: A 768× 640 pixels 200Meps dynamic vision sensor," in *IEEE International Symposium on Circuits and Systems (ISCAS)*, Baltimore, MD, USA, 2017.

[13] A. Z. Zhu, D. Thakur, T. Ozaslan, B. Pfrommer, V. Kumar and K. Daniilidis, "The Multi Vehicle Stereo Event Camera Dataset: An Event Camera Dataset for 3D Perception," *IEEE Robotics and Automation Letters,* vol. 3, no. 3, pp. 2032-2039, 2018. DOI: 10.1109/LRA.2018.2800793.

[14] E. Mueggler, H. Rebecq, G. Gallego, T. Delbruck and D. Scaramuzza, "The event-camera dataset and simulator: Event-based data for pose estimation, visual odometry, and SLAM," *The International Journal of Robotics Research,* vol. 36, no. 2, pp. 142-149, 2017.

[15] H. Rebecq, D. Gehrig and D. Scaramuzza, "Esim: an open event camera simulator," in *Conference on Robot Learning* , Zürich, Switzerland, 2018.

[16] J. A. Leñero-Bardallo, R. Carmona-Galán and A. Rodríguez-Vázquez, "Applications of event-based image sensors—Review and analysis," *International Journal of Circuit Theory and Applications,* vol. 46, no. 9, pp. 1620-1630, 2018.

[17] N. Wu, "Neuromorphic vision chips," *Science China Information Sciences,* 2018. DOI: 10.1007/s11432-017-9303-0.

[18] P. Rogister, R. Benosman, S.-H. Ieng, P. Lichtsteiner and T. Delbruck, "Asynchronous Event-Based Binocular Stereo Matching," *IEEE Transactions on Neural Networks and Learning Systems* , vol. 23, no. 2, pp. 347 - 353, 2011. DOI: 10.1109/TNNLS.2011.2180025.

[19] J. Carneiro, S.-H. Ieng, C. Posch and R. Benosman, "Event-based 3D reconstruction from neuromorphic retinas," *Neural Networks,* vol. 45, pp. 23-38, 2013.

[20] Z. Xie, S. Chen and G. Orchard, "Event-based stereo depth estimation using belief propagation," *Frontiers in neuroscience,* vol. 11:535, 2017. DOI: 10.3389/fnins.2017.00535.

[21] J. Kogler, C. Sulzbachner and W. Kubinger, "Bio-inspired Stereo Vision System with Silicon Retina Imagers," *Computer Vision Systems. ICVS 2009. Lecture Notes in Computer Science,* vol. 5815, pp. 174-183, 2009. DOI: 10.1007/978-3-642-04667-4_18.

[22] S. Schraml, A. N. Belbachir, N. Milosevic and P. Schön, "Dynamic stereo vision system for real-time tracking," in *IEEE International Symposium on Circuits and Systems*, Paris, France, 2010.

[23] L. A. Camuñas-Mesa, T. Serrano-Gotarredona, S. H. Ieng, R. B. Benosman and B. Linares-Barranco, "On the use of orientation filters for 3D reconstruction in event-driven stereo vision," *Frontiers in Neuroscience,* vol. 8, no. 48, 2014, DOI: 10.3389/fnins.2014.00048.

[24] A. Linares-Barranco, F. Gomez-Rodriguez, A. Jimenez-Fernandez, T. Delbruck and P. Lichtensteiner, "Using FPGA for visuo-motor control with a silicon retina and a humanoid robot," in *IEEE International Symposium on Circuits and Systems*, New Orleans, LA, USA, 2007. DOI: 10.1109/ISCAS.2007.378265 .

[25] S. Hoseini and B. Linares-Barranco, "Real-Time Temporal Frequency Detection in FPGA Using Event-Based Vision

Sensor," in *IEEE 14th International Conference on Intelligent Computer Communication and Processing (ICCP)*, Cluj-Napoca, Romania, 2018. DOI: 10.1109/ICCP.2018.8516629.

[26] M. Liu and T. Delbruck, "Block-matching optical flow for dynamic vision sensors: Algorithm and FPGA implementation.," in *IEEE International Symposium on Circuits and Systems (ISCAS)*, Baltimore, MD, USA, 2017. DOI: 10.1109/ISCAS.2017.8050295.

[27] I. Shour, "A reconfigurable architecture for event-based optical flow in FPGA," MPhil. Thesis. Politecnico di Torino., 2018.

[28] F. Eibensteiner, J. Kogler and J. Scharinger, "A high-performance hardware architecture for a frameless stereo vision algorithm implemented on a FPGA platform," in *Proceedings of the IEEE Conference on Computer Vision and Pattern Recognition Worksh*, Columbus, OH, USA, 2014.

[29] "JMonkey Engine," [Online]. Available: http://jmonkeyengine.org/.

[30] H. Simmler, L. Levinson and R. Männer, "Multitasking on FPGA coprocessors," *Lecture Notes in Computer Science,* vol. 1896, pp. 121-130, 2000.

[31] C. Kachris and D. Soudris, "A survey on reconfigurable accelerators for cloud computing," in *26th International Conference on Field Programmable Logic and Applications (FPL)*, Lausanne, Switzerland, 2016.

[32] H. M. Waidyasooriya, M. Hariyama and K. Uchiyama, Design of FPGA-Based Computing Systems with OpenCL, Cham, Switzerland: Springer International Publishing, 2018.

[33] K. Kang and P. Yiannacouras, "Host Pipes: Direct Streaming Interface Between OpenCL Host," in *Proceedings of the 5th International Workshop on OpenCL (IWOCL 2017)*, Toronto, Canada, 2017. DOI: 10.1145/3078155.3078182.

[34] R. Kobayashi, Y. Oobata, N. Fujita, Y. Yamaguchi and T. Boku, "OpenCL-ready high speed FPGA network for reconfigurable high performance computing," in *Proceedings of the International Conference on High Performance Computing in Asia-Pacific Region*, Tokyo, Japan, 2018.

[35] K. Hill, S. Craciun, A. George and H. Lam, "Comparative analysis of OpenCL vs. HDL with image-processing kernels on Stratix-V FPGA," in *IEEE 26th International Conference on Application-specific Systems, Architectures and Processors (ASAP)*, Toronto, Canada, 2015.

[36] H. Rebecq, G. Gallego and D. Scaramuzza, "EMVS: Event-based Multi-View Stereo," in *British Machine Vision Conference (BMVC)*, York, UK, 2016.

[37] A. Z. Zhu, Y. Chen and K. Daniilidis, "Realtime Time Synchronized Event-based Stereo," *arXiv preprint arXiv:,* vol. 1803.09025, 2018.